# A Study of Car-to-Train Assignment Problem for Rail Express Cargos on Scheduled and Unscheduled Train Service Network

Boliang Lin[*]

School of Traffic and Transportation, Beijing Jiaotong University, Beijing 100044, China

**Abstract:** Freight train services in a railway network system are generally divided into two categories: one is the unscheduled train, whose operating frequency fluctuates with origin-destination (OD) demands; the other is the scheduled train, which is running based on regular timetable just like the passenger trains. The timetable will be released to the public if determined and it would not be influenced by OD demands. Typically, the total capacity of scheduled trains can usually satisfy the predicted demands of express cargos in average. However, the demands are changing in practice. Therefore, how to distribute the shipments between different stations to unscheduled and scheduled train services has become an important research field in railway transportation. This paper focuses on the coordinated optimization of the rail express cargos distribution in two service networks. On the premise of fully utilizing the capacity of scheduled service network first, we established a Car-to-Train (CTT) assignment model to assign rail express cargos to scheduled and unscheduled trains scientifically. The objective function is to maximize the net income of transporting the rail express cargos. The constraints include the capacity restriction on the service arcs, flow balance constraints, logical relationship constraint between two groups of decision variables and the due date constraint. The last constraint is to ensure that the total transportation time of a shipment would not be longer than its predefined due date. Finally, we discuss the linearization techniques to simplify the model proposed in this paper, which make it possible for obtaining global optimal solution by using the commercial software.

---

[*] Corresponding author. Email: bllin@bjtu.edu.cn

# 1. Introduction

In recent years, the competition between railway and highway transportation is more and more intense, especially for the long-distance transportation of high-value freight. The governments expect that more freight flow on highway should be diverted to railway in order to reduce carbon emissions. For example, in Europe, 30% of road freight over 300 km is expected to shift to rail or waterborne transport by 2030. The Ministry of Transport of China also recommended that railway and waterway should undertake more freight transportation. However, road freight transport is convenient and efficient nowadays, and many shippers tend to choose road transport. Therefore, as one of the largest and busiest railway systems, China railway paid much attention to the development of scheduled train service in past years, wishing to get more market share on the high-value freight transportation. For example, on the official website of the China Railway, we can find 156 scheduled train services, including trains which are bound for the West-Asia and the Europe. The speed of scheduled trains in China can be divided into the following three levels, 160 km/h, 120 km/h and 80 km/h (the speed of regular freight train is usually less than 50km/h). The scheduled train plan is usually made on the basis of historical data and the experience of the staffs. However, some of the train services attract less freight flows than expected. In practice, the unreasonable plan can be improved only when the failure has occurred. Based on our previous estimation, the transportation demands of high-value freight is very large and is far beyond the supply of the 156 train services. Therefore, in this situation, it is necessary to improve the railway freight transport service to attract more freight flow from road.

The problem is whether the market-survey-based scheduled train service plan provided by the railway company is reasonable. If a simulation can be conducted before the application of train service plan, the decision failure can be reduced significantly. Therefore, optimizing the car-to-train assignment for the rail express cargo is an important problem that needs to be addressed both theoretically and practically.

The remainder of this paper is organized as follows. Section 2 presents works related to the car-to-block assignment and the block-to-train assignment problem. Section 3 describes the Car-to-Train assignment problem of rail express cargo in detail. Section 4 provides a model formulation and linearization techniques of it. Conclusions are presented in Section 5.

## 2. Related Work

Generally, the core of the train connection service problem is involved with two consolidation processes: the car-to-block assignment and the block-to-train assignment.

The objective of the car-to-block assignment problem is to determine which blocks should be built at each yard and the assignment of cars to these blocks. The objective is to minimize the total cost. One of the first models is proposed by Bodin et al. [1], who developed a nonlinear model to determine a classification plan for all the classification yards, which can be viewed as a multi-commodity flow problem with capacity constraints in terms of the maximum number of blocks and the maximum car volume that can be handled. In recent researches, the blocking problem is regarded as a very large-scale, multi-commodity, service network design problem with billions of decision variables. Different heuristic optimization algorithms are proposed to solve the problem. Barnhart et al. [2] formulated the railroad blocking problems as a network design problem, considered the flow constraints on the nodes and arcs, proposed Lagrangian relaxation heuristic algorithm to solve the problem. The model is tested on a major railroad, and the validity of the model and algorithm are verified. Gorman [3] addressed the joint train-scheduling and demand-flow problem for a major US freight railroad. The tabu-enhanced genetic search is used to find acceptable solutions, which consistently achieves better approximations to the optimum and maintains its performance as the problem size grows. Ahuja et al. [4] developed the very large-scale neighborhood (VLSN) search algorithm to solve the blocking problem, which is able to get the solution to near optimality using one to two hours of computer time on a standard workstation computer.

The block-to-train problem is to determine which train services are to be supplied at what frequency and the assignment of blocks to train services. Thomet [5] developed a cancellation procedure that gradually replaces direct trains with a series of intermediate train connections in order to minimize operation and delay costs. Kwona et al. [6] used a time-space network technique to improve a given blocking plan and the block-to-train assignment. The problem formulated as a linear multi-commodity flow problem and the column generation technique was used as a solution approach.

Since the above two sub-problems are interrelated, some researchers consider these two issues as an integrated problem, establish a car-to-train model directly to solve train connection service problem at once. Keaton [7] formulated the combined problem as a 0-1 mixed integer programming model which is to minimize the sum of train costs, car time costs, and classification yard costs, while not exceeding limits on train size and yard volumes. A heuristic approach based on Lagrangian relaxation was presented to solve the problem. Crainic et al. [8] formulated a general optimization model which takes into account the problem of routing of traffic and scheduling train services. A heuristic algorithm was developed to solve this mixed-integer multi-commodity problem.

## 3. Problem Descriptions

For a railway service network, it may consist of multiple types of train services, such as the local train services, the through train services and the express train services. Therefore, to transport a given shipment from its origin to its destination, we can use either a single type of train service or a combination of multiple train services. To have a better understanding of the overall transportation process, we would like to give more details using the example depicted in Figure 1.

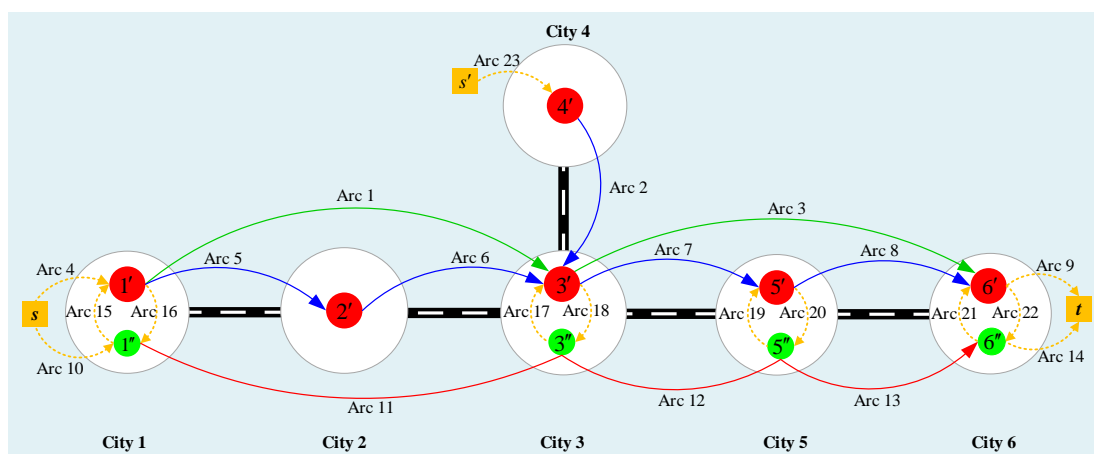

**Figure 1.** The railway service network.

In Figure 1, the red circles $1' \sim 6'$ represent railway classification yards, the green circles $1'', 3'', 5'', 6''$ are railway logistics centers, the golden diamonds $s, s'$ and $t$ denote big enterprises that have direct accesses (i.e. enterprises' special rail lines) to railways. The blue arcs Arc 2, Arc 5 ~ Arc 8 represent shuttle train services (formed at one classification yard and broken up at the adjacent yard), and the green arcs Arc 1 and Arc 3 are through train services (i.e. long-distance ordinary train services that are formed at one classification yard, pass through one or more yards, and finally are broken up at a relatively far yard). Moreover, the yellow dotted arcs Arc 4, Arc 9, Arc 10, Arc 14 ~ Arc 23 are local train services (or pickup and delivery train services between adjacent yards that are powered by shunting engines), and the red arcs Arc 11, Arc 12 and Arc 13 represent express train service arcs, which belong to the same express train service $1'' \to 6''$. Since the express train service $1'' \to 6''$ have car block swap operations at logistics centers $3''$ and $5''$, we decompose the train services into three connected service arcs.

Consider the two shipments $s \to t$ and $s' \to t$, and their required transportation due date are $T_{st}$ and $T_{s't}$, respectively. Now let us analyze the possible transportation strategies for the first shipment $s \to t$. One possible strategy is that the shipment is first transported to yard $1'$ by the local train service Arc 4 from its origin $s$, and is then transported to yard $3'$ carried by the through train service Arc 1. After the classification operation, the shipment is grouped into the through train service Arc 3 and transported to the last yard $6'$ on its itinerary. Finally, the shipment is sent to its destination $t$ by the local train service Arc 9 to complete its entire itinerary. During the overall transportation process, the shipment get reclassified three times (at yards $1'$, $3'$ and $6'$). The train service chain used by the train service is as follows:

(1) $s \to \text{Arc4} \to \text{Arc1} \to \text{Arc3} \to \text{Arc9} \to t$.

Other possible transportation strategies that can be adopted by the shipment are as follows (but are not limited to):

(2) $s \to \text{Arc4} \to \text{Arc1} \to \text{Arc7} \to \text{Arc8} \to \text{Arc9} \to t$;

(3) $s \to \text{Arc4} \to \text{Arc5} \to \text{Arc6} \to \text{Arc7} \to \text{Arc8} \to \text{Arc9} \to t$;

(4) $s \to \text{Arc}10 \to \text{Arc}11 \to \text{Arc}12 \to \text{Arc}13 \to \text{Arc}14 \to t$.

In Strategy (2), the shipment needs to be classified four times (at yards $1'$, $3'$, $5'$ and $6'$). While in Strategy (3), the shipment needs to be classified five times (at yards $1'$, $2'$, $3'$, $5'$ and $6'$). In contrast, only two times of classification operations (at yards $1'$ and $6'$) are needed in Strategy (4). However, delays are occurred due to the car block swap operations at logistics centers $3''$ and $5''$ in the strategy.

Similarly, for another shipment $s' \to t$, the following transportation strategies can be adopted:

(1) $s' \to \text{Arc}23 \to \text{Arc}2 \to \text{Arc}3 \to \text{Arc}9 \to t$;

(2) $s' \to \text{Arc}23 \to \text{Arc}2 \to \text{Arc}7 \to \text{Arc}8 \to \text{Arc}9 \to t$;

(3) $s' \to \text{Arc}23 \to \text{Arc}2 \to \text{Arc}18 \to \text{Arc}12 \to \text{Arc}13 \to \text{Arc}14 \to t$.

In Strategy (1) and (2), the shipment gets reclassified three times and four times, respectively. In contrast, in Strategy (3), the shipment gets reclassified three times and has one time of car block swap operations.

The train services are characterized by costs, operating speed, capacity and service frequency. For example, the express train service is usually of high costs, fast speed and low frequency; while the district train service has lower speed at a lower cost, and is always provided more frequently. As a result, some of the transportation strategies could violate the due date restrictions. Furthermore, some other strategies possibly result in overloads on the links through which the consider shipments and other shipments pass together. For instance, Strategy (1) of transporting shipment $s \to t$ and Strategy (1) of transporting shipment $s' \to t$ use the common through train service, i.e. Arc 3. If the volumes of these two shipments are relatively high, due to limited capacity on Arc 3, its capacity constraint may get violated when above two strategies are adopted simultaneously. Therefore, how to select optimal transportation strategies for all the shipments respecting the due date constraints and capacity constraints is a typically complicated combinatorial optimization problem.

## 4. Mathematical Model

This section aims to provide a mathematical description for the rail express cargo car-to-train assignment problem. To facilitate the model formulation, we make the following assumptions throughout this paper:

(1) Each city has at least one railway station, and different stations (including marshalling stations and logistics centers) in a city can be connected by local train services.

(2) Arcs between any two nodes are regarded as different arcs even they belong to the same train services. For example, in Figure 1, though Arc 11 and Arc 12 belong to the same express train services, they are treated as two different arcs. Note that these two arcs are connected in Logistics Center $3''$, where car block swap operations may take place.

(3) Each shipment should not be split during the transportation process. This means that each shipment can only choose a single route (a train service chain). However, railway operators can decide whether to transport the whole volume of a shipment or just a portion of it.

(4) Considering the fact that the frequency of regular freight train fluctuates with the traffic volume (once the volume reaches the predefined size of a train, a freight train will be dispatched. Thus, the number of trains dispatched per day is fluctuant). As the traffic volume of each regular train arc is much larger than the express train arc, it is assumed that the traffic flows exceeding the capacity of an express train arc can be definitely transported by a regular train arc.

### 4.1 Notations

The notations used in the mathematical model are described as follows:

(1) Sets

$V$ : Set of all nodes, yards and logistics centers;

$E$ : Set of train service arcs;

$E^{\text{express}}$ : Set of express train service arcs;

$G$ : Set of shipments, $g \in G$;

(2) Parameters

$s_m$ : Start node of arc $m$, $s_m \in V$, $m \in E$;

$t_m$ : Tail node of arc $m$, $t_m \in V$, $m \in E$;

$C_m$ : Capacity of arc $m$;

$L_m$ : Length of arc $m$;

$\tau_m$ : Transportation time on arc $m$;

$\tau_{mn}$ : Transfer time from arc $m$ to arc $n$ for a car.

$N_g$ : Volume of the $g$th shipment that originates at $i$ and destinates to $j$, $i \in V_{supply}$, $j \in V_{demand}$;

$o_g$ : The origin of the $g$th shipment;

$d_g$ : The destination of the $g$th shipment;

$R_g$ : Per car cost of transporting the $g$th shipment paid by the shippers;

$T_g$ : Transportation due date of the $g$th shipment;

$\lambda$ : the cost of express train service per car kilometer;

(3) Decision variables

$\xi_g$ : Continuous variables. Proportion of volume of the $g$th shipment that are able to be transported;

$x_g^m$ : Binary variables. $x_g^m = 1$ if the $g$th shipment uses the $m$th service arc; $x_g^m = 0$ otherwise.

## 4.2 Model Formulation

According to the notations above, the CTT assignment problem for the rail express cargo can be written as follows:

Objective function:

$$\max \sum_{g \in G} R_g N_g \xi_g - \lambda \sum_{m \in E^{\text{express}}} C_m L_m \tag{1}$$

Subject to

$$\sum_{m \in E} x_g^m \tau_m + \sum_{m \in E} \sum_{n: s_n = t_m} \tau_{mn} x_g^m x_g^n \leq T_g \quad \forall g \in G \tag{2}$$

$$\sum_{g \in G} N_g \xi_g x_g^m \leq C_m \quad \forall m \in E \tag{3}$$

$$M \xi_g \geq x_g^m \quad \forall g \in G, m \in E \tag{4}$$

$$\xi_g \sum_{m: s_m = o_g} x_g^m = \xi_g, \quad \forall g \in G \tag{5}$$

$$\xi_g \sum_{m: t_m = d_g} x_g^m = \xi_g, \quad \forall g \in G \tag{6}$$

$$\xi_g \sum_{m: t_m = k} x_g^m = \xi_g \sum_{n: s_n = k} x_g^n, \quad \forall g \in G, k \in V, k \neq o_g, k \neq d_g \tag{7}$$

$$x_g^m \in \{0,1\} \quad \forall g \in G, m \in E \tag{8}$$

$$\xi_g \in [0,1] \quad \forall g \in G \tag{9}$$

In the model above, Eq. (1) is the objective function, which maximizes the net income of transporting the rail express cargo. The net income is equal to the total transportation fees paid by the shippers minus the total operating costs of organizing the express train services by the railway company.

The second term in the objective function is the total operation cost of all express trains provided by Railway Company. Set the fixed operation cost of express train $i$ is $C_i^{\text{fix}}$, the cost of per travelling mileage is $C_i^{\text{TrainRun}}$ and the transport distance is $\ell_i$. They meet the following equation:

$$\lambda \sum_{m \in E^{\text{express}}} C_m L_m = \sum_{i \in S^{\text{express}}} \left( C_i^{\text{fix}} + C_i^{\text{TrainRun}} \ell_i \right) \tag{10}$$

In formula (10), $S_i^{\text{express}}$ is the set of express trains.

It should be noted that express trains are different from regular trains which can be dispatched once the arrival car volume reaches the predefined train size. The operation frequency of express train is fixed in a planning horizon. In other words, an express train should also be dispatched on schedule, even if the car volume is less than the given threshold. Therefore, the second term in formula (10) is a constant which is unnecessary to be optimized.

Since the service network is given in advance, the latter costs are a constant, which means it does not affect the optimal solution. Therefore, in the computational experiments, we will remove the constant express train service organization costs from the objective function.

Constraint (2) ensures that the total transportation time of a shipment should not be longer than its predefined due date. In this study, the total transportation time consists of the transportation time on arcs and the transfer time between arcs. Constraint (3) is the capacity restriction on the service arcs. Constraint (4) guarantees the logical relationship between the two groups of decision variables. Constraint (5) ~ (7) are the well-known flow balance constraints. Specifically, Constraint (5) is flow balance constraint at departure nodes; Constraint (6) is flow balance constraint at arrival nodes; and Constraint (7) is flow balance constraint at intermediate (passing) nodes. Finally, decision variable domains are specified by Constraints (8) and (9).

## 4.3 Linearization of the Transportation Due Date Constraint

Clearly, due date constraint (2) is a non-linear constraint because it involves the production of two decision variables. In general, a linear model is easier to solve than a non-linear one. We are hence motivated to linearize original constraint (2) to a linear constraint. To this end, we need to first introduce an auxiliary binary decision variable $y_{stg}^{mn}$ and it is defined as follows:

$$y_g^{mn} = x_g^m \cdot x_g^n \qquad \forall g \in G, m \in E, n \in E, t_m = s_n \tag{11}$$

In this way, constraint (2) can be converted to:

$$\sum_{m \in E} x_g^m t_m + \sum_{m \in E} \sum_{n: s_n = t_m} \tau_{mn} y_g^{mn} \leq T_g \quad \forall g \in G \tag{12}$$

Besides, we add the following additional constraints into the original model:

$$x_g^m + x_g^n - 1 \leq y_g^{mn} \leq \frac{1}{2} \cdot \left( x_g^m + x_g^n \right) \quad \forall g \in G, m \in E, n \in E, t_m = s_n \tag{13}$$

$$y_g^{mn} \in \{0,1\} \qquad \forall g \in G, m \in E, n \in E, t_m = s_n \tag{14}$$

By replacing $x_g^m \cdot x_g^n$ with $y_g^{mn}$ in constraint (2) and by adding constraint (13) and (14) into the original model, we are managed to transform constraint (3) to a linear constraint, which means we can directly use a standard optimization solver (e.g. CPLEX or Gurobi) to solve the resulting model.

## 5. Conclusions

In this paper, a car-to-train assignment model for rail express cargos is established. In light of the characteristics of high value-added goods, the due date constraint is added to the model, considering both transportation time on arcs and transfer time between arcs. Besides, the arc capacity constraint and flow balance constraint are also considered. Moreover, linearization techniques are used to simplify the model in order to reduce the difficulty in solving the model.

Note that, it is assumed that the train operation plan is given, which is developed mainly relied on manual process in practice, without any optimization-based approach. In this case, the resulting car-to-train assignment plan can serve as a solid aid in improving the quality of train operation plan. In the long term, researchers can focus on the joint optimization of train operation plan and car-to-train assignment plan.